\documentclass[sigconf,anonymous=false]{acmart}
\usepackage{latexsym}
\usepackage{marvosym}
\usepackage{tikz}
\usepackage{amsmath}
\usepackage{amsfonts}
\usepackage{booktabs}
\usepackage{tabularx}
\usepackage{caption}
\usepackage{subcaption}
\usepackage{filecontents}
\usepackage{multirow}
\usepackage{enumitem}
\usepackage{url}

\def\BibTeX{{\rm B\kern-.05em{\sc i\kern-.025em b}\kern-.08emT\kern-.1667em\lower.7ex\hbox{E}\kern-.125emX}}
    
\begin{document}
%
\title{Holmes: Towards Distributed Training Across Clusters with Heterogeneous NIC Environment}



\author{Fei Yang}
\affiliation{
  \institution{Zhejiang Lab}
  \country{China}}
\email{yangf@zhejianglab.com}

\author{Shuang Peng}
\affiliation{
  \institution{Zhejiang Lab}
  \country{China}}
\email{pengs@zhejianglab.com}

\author{Ning Sun}
\affiliation{
  \institution{Zhejiang Lab}
  \country{China}}
\email{sunning@zhejianglab.com}

\author{Fangyu Wang}
\affiliation{
  \institution{Zhejiang Lab}
  \country{China}}
\email{wangfy@zhejianglab.com}

\author{Yuanyuan Wang}
\affiliation{
  \institution{Zhejiang Lab}
  \country{China}}
\email{wangyy2022@zhejianglab.com}

\author{Fu Wu}
\affiliation{
  \institution{Zhejiang Lab}
  \country{China}}
\email{wufu@zhejianglab.com}

\author{Jiezhong Qiu}
\affiliation{
  \institution{Zhejiang Lab}
  \country{China}}
\email{jiezhongqiu@zhejianglab.com}

\author{Aimin Pan*}
\affiliation{
  \institution{Zhejiang Lab}
  \country{China}}
\email{panaimin@zhejianglab.com}

\thanks{* The corresponding author.}

\renewcommand{\shortauthors}{Yang et al.}

%
\begin{abstract}
Large language models (LLMs) such as GPT-3, OPT, and LLaMA have demonstrated remarkable accuracy in a wide range of tasks. However, training these models can incur significant expenses, often requiring tens of thousands of GPUs for months of continuous operation. Typically, this training is carried out in specialized GPU clusters equipped with homogeneous high-speed Remote Direct Memory Access (RDMA) network interface cards (NICs). The acquisition and maintenance of such dedicated clusters is challenging. Current LLM training frameworks, like Megatron-LM and Megatron-DeepSpeed, focus primarily on optimizing training within homogeneous cluster settings. In this paper, we introduce Holmes, a training framework for LLMs that employs thoughtfully crafted data and model parallelism strategies over the heterogeneous NIC environment. Our primary technical contribution lies in a novel scheduling method that intelligently allocates distinct computational tasklets in LLM training to specific groups of GPU devices based on the characteristics of their connected NICs. Furthermore, our proposed framework, utilizing pipeline parallel techniques, demonstrates scalability to multiple GPU clusters, even in scenarios without high-speed interconnects between nodes in distinct clusters. We conducted comprehensive experiments that involved various scenarios in the heterogeneous NIC environment. In most cases, our framework achieves performance levels close to those achievable with homogeneous RDMA-capable networks (InfiniBand or RoCE), significantly exceeding training efficiency within the pure Ethernet environment. Additionally, we verified that our framework outperforms other mainstream LLM frameworks under heterogeneous NIC environment in terms of training efficiency and can be seamlessly integrated with them. 
\end{abstract}


%


%
\maketitle

\section{Introduction}
In recent years, the emergence of large language models (LLMs)~\cite{smith2022using,zhao2023survey} such as GPT-3~\cite{brown2020language}, OPT~\cite{zhang2022opt}, and LLaMA~\cite{touvron2023llama} has significantly transformed the landscape of natural language processing (NLP) by achieving state-of-the-art accuracy in a wide variety of tasks. Despite these remarkable advancements, LLM training is often accompanied by considerable costs, requiring the sustained operation of tens of thousands of advanced GPUs over extended periods. For instance, Meta's 175 billion parameter OPT-175 was trained for 33 days on 1024 NVIDIA A100 GPUs. A slightly smaller model, LLaMa with 65 billion parameters, was trained for 21 days using a cluster of 2048 NVIDIA A100 GPUs~\cite{zhang2022opt,touvron2023llama, brown2020language,yuan2022decentralized}.
Conventionally, these models undergo training within specialized clusters that are equipped with homogeneous high-speed Remote Direct Memory Access (RDMA) network interface cards (NICs). However, the acquisition and maintenance of such dedicated clusters can present substantial financial and logistical challenges~\cite{touvron2023llama,zhao2023survey}.

\begin{figure*}[t]
  \centering
  \includegraphics[width = 0.84\textwidth]{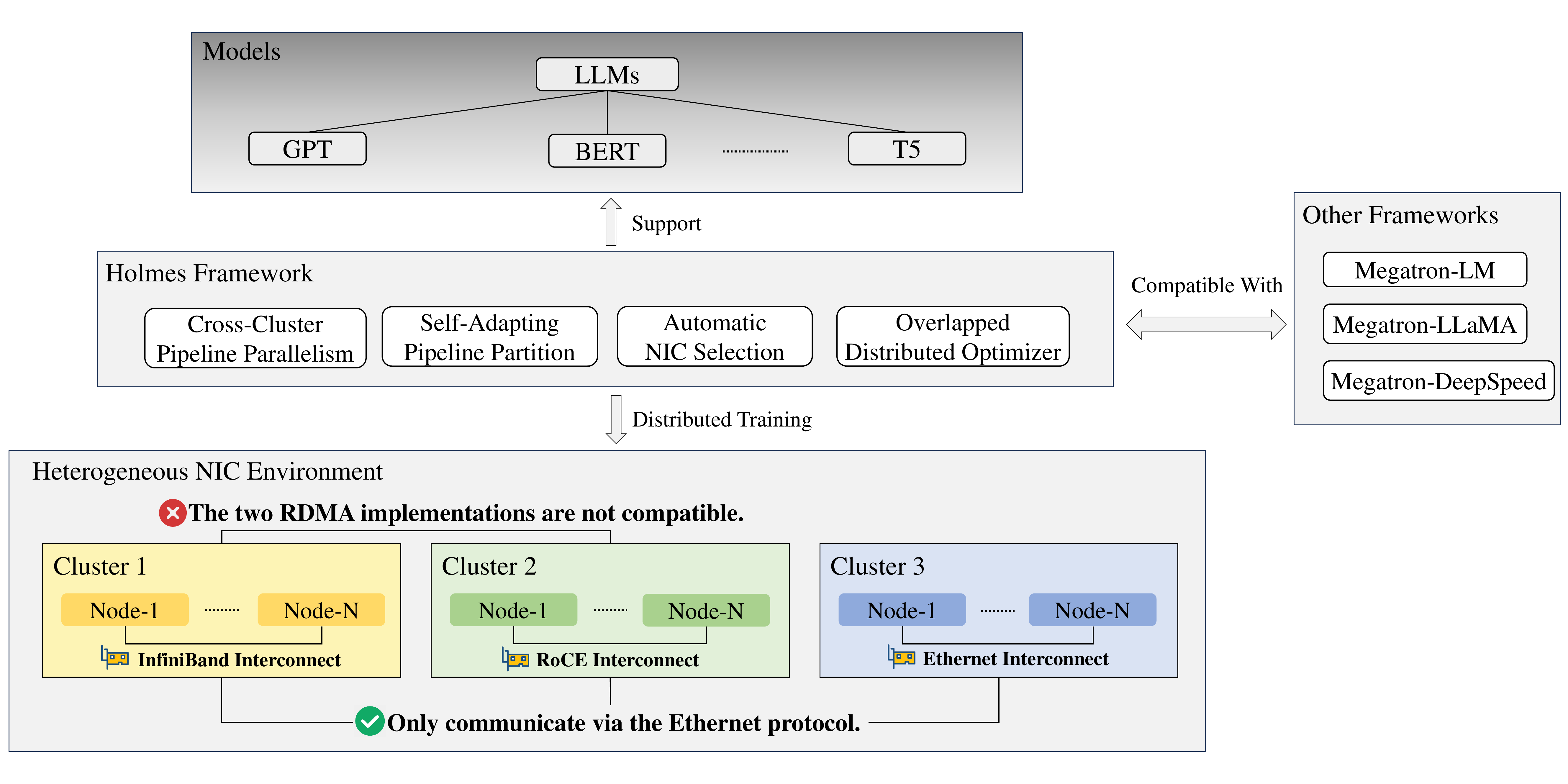}
  \caption{An overview of the Holmes framework. Holmes provides training support for a diverse range of LLM types and seamlessly integrates with contemporary mainstream LLM training frameworks. This flexibility extends to the training of LLMs in the heterogeneous NIC environment, where Holmes consistently achieves performance levels that are close to those achievable in the homogeneous NIC settings.}
  \label{frame}
\end{figure*}
RDMA technology has found extensive deployment in modern data centers~\cite{Flor,gao2021cloud}, offering low latency and high throughput benefits that are particularly advantageous for LLM training. RDMA technology primarily comprises two categories: InfiniBand and RDMA over Converged Ethernet (RoCE)~\cite{chen2016fast,guo2016rdma,gao2021cloud}. Both of these technologies facilitate direct access to remote computer memory, bypassing the need for CPU intervention, thus leading to more efficient data transfer and communication. InfiniBand represents a dedicated networking technology, while RoCE is a protocol designed to implement RDMA within the Ethernet framework. Both InfiniBand and RoCE play an integral role in the construction of high-performance computing and data center networks~\cite{kalia2016design,guo2016rdma,gao2021cloud}.
Despite their widespread applications, the two networking technologies, InfiniBand and RoCE, are inherently incompatible with each other. This incompatibility poses challenges, especially in the context of training LLMs within or across clusters that incorporate various RDMA NICs. In such scenarios, the adoption of pure Ethernet for communication often emerges as the only viable option.

Speeding up the LLM training process and minimizing associated costs have been persistent areas of research. Current LLM training frameworks, as evidenced in recent surveys~\cite{zhao2023survey}, which feature prominent models like Megatron-LM~\cite{shoeybi2019megatron,narayanan2021efficient}, DeepSpeed~\cite{rasley2020deepspeed,smith2022using,deepspeed}, and FairScale~\cite{FairScale2021}, predominantly prioritize speed within homogeneous data center networks. Consequently, these frameworks may not be easily adaptable to the diverse landscape of the heterogeneous RDMA NIC environment. To address these challenges, there is a need for an open and unified LLM training framework that can seamlessly accommodate the heterogeneity of RDMA NICs across various clusters, thus alleviating the operational complexity associated with large-scale data center networks.

In this paper, we present Holmes, a high-performance framework optimized for training LLMs in the heterogeneous RDMA NIC environment. An overview of the framework is illustrated in Figure~\ref{frame}. Holmes optimally leverages the diverse capabilities of NICs within cluster configurations. A key contribution lies in our intelligent scheduling methodology, strategically allocating computational tasklets to GPUs based on their associated NIC types. This approach aligns computational tasklets with the strengths of each network technology, including InfiniBand, RoCE, and Ethernet. As a result, Holmes significantly mitigates overall training costs.

Specifically, our framework tracks hardware topology and network characteristics, dynamically assigning parallel work groups to suitable GPU devices during LLM training. The internal scheduling method, which is applicable to the heterogeneous NIC environment, is complemented by well-designed optimizations in both data and model parallelism~\cite{xing2015petuum,huang2019gpipe,narayanan2019pipedream,li2020pytorch}. As a result, Holmes can effectively harness the full potential of the real-world production infrastructure. This includes the use of high-speed RDMA NICs for data parallelism, which exhibits a high communication overhead~\cite{li2020pytorch,rajbhandari2020zero,ren2021zero}, and Ethernet NICs for pipeline model parallelism, known for its lower communication overhead~\cite{huang2019gpipe}. We demonstrate impressive performance in using our framework to train LLM across clusters equipped with heterogeneous NICs.

The proposed Holmes framework demonstrates exceptional scalability through pipeline model parallelism, seamlessly extending to multiple GPU clusters. This scalability is maintained even in scenarios where high-speed interconnects (InfiniBand or RoCE) are unavailable between nodes from distinct clusters. Therefore, Holmes can be applied to train LLMs across multiple GPU clusters at different locations. This maximizes the utilization of existing GPU clusters for LLM training, eliminating the need to reconstruct larger GPU clusters to accommodate numerous GPU devices and RDMA NICs specifically for training purposes. To validate Holmes's scalability, we conducted an extensive range of experiments covering a diverse spectrum of scenarios involving heterogeneous NICs. Remarkably, our training framework consistently achieved performance levels close to those attainable in environments with homogeneous RDMA-capable networks, such as pure InfiniBand or RoCE. Furthermore, Holmes notably surpassed training efficiency in the pure Ethernet environment.


\begin{figure*}[t]
  \centering
  \includegraphics[width = 0.84\textwidth]{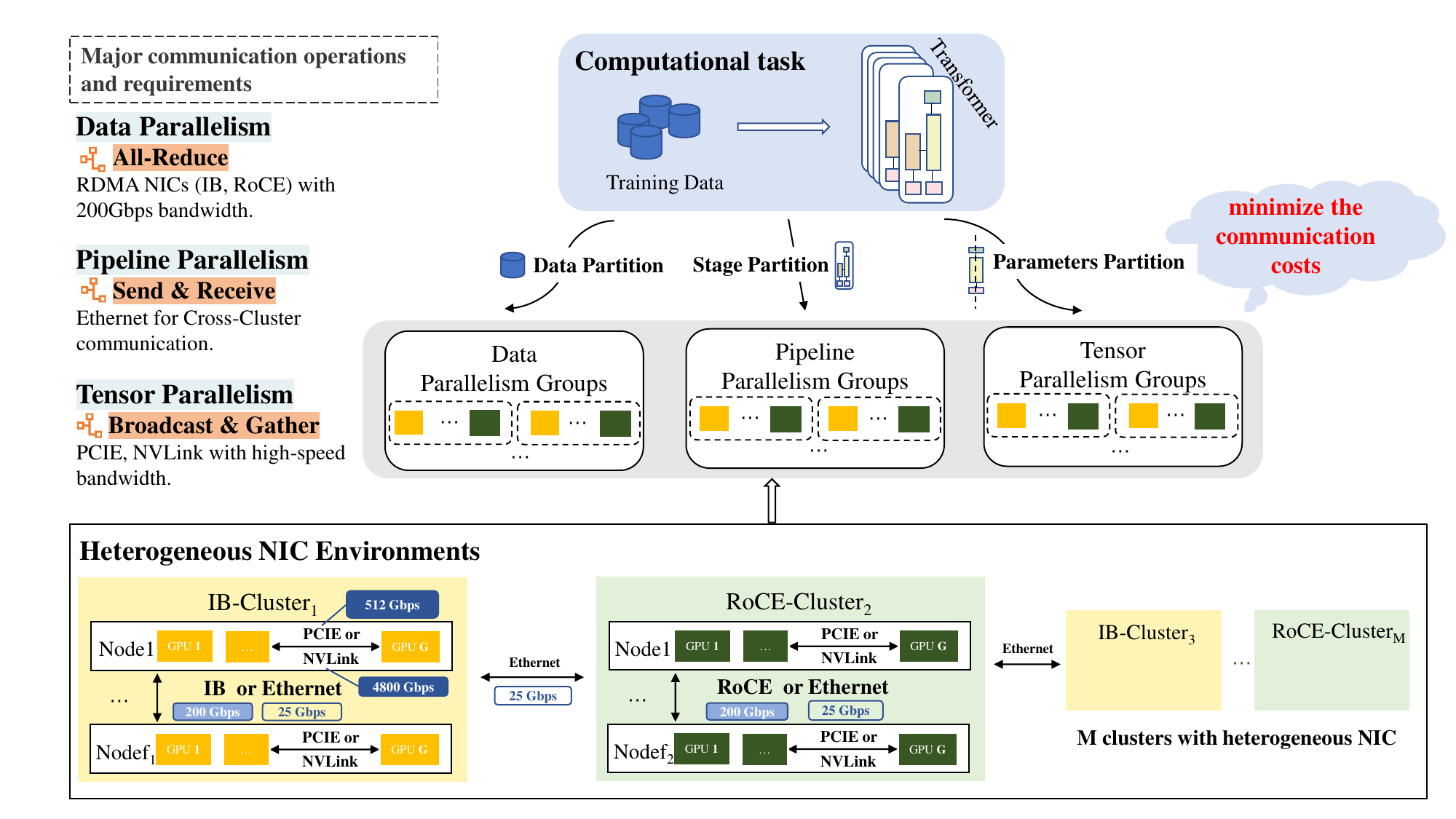}
  \caption{An illustration of the scheduling problem in heterogeneous NIC environments, with the objective of minimizing communication costs for distributed LLM training tasks.}
  \label{problem description}
\end{figure*}

In general, we firmly believe that Holmes signifies a notable advancement in democratizing LLMs, making efficient scaling attainable for a broader research community. 
Our contributions are summarized as follows:
\begin{itemize}[itemsep=1pt,topsep=1pt,parsep=1pt,leftmargin=0.4cm]
 \item We propose Holmes, a novel LLM training framework designed to adapt to the heterogeneous NIC environment. Holmes is non-intrusive to existing infrastructures, harnessing data and model parallelism technologies. It facilitates LLM training across multiple GPU clusters without the need to reconstruct larger GPU clusters. This feature is crucial for conserving computational resources.
 \item Moreover, Holmes introduces a method for tailoring stage division in pipeline model parallelism. This method takes into account the speed variations inherent in heterogeneous NICs and allocates varying numbers of layers to different stages accordingly.
 \item We conducted comprehensive experiments to demonstrate that training LLMs with Holmes can achieve performance close to those achievable with homogeneous RDMA-capable networks, significantly exceeding training efficiency within the pure Ethernet environment.
 \item Furthermore, Holmes seamlessly integrates with other mainstream LLM training frameworks such as Megatron-LM and Megatron-DeepSpeed.
\end{itemize}

\textbf{Limitations and Future Directions.} In this paper, we focus on a fundamental part of LLM training, but there are important practical aspects that we have not yet covered. We assume that communication between devices is stable and that all devices are consistently online. It is important to note that we have not fully trained the system to convergence; instead, we have performed partial training to validate our approach. In the future, we need to explore scheduling methods for diverse environments and figure out how to handle faults.

\section{Preliminary}
In this section, we first provide an overview of LLM infrastructure. Following this, we introduce different NIC environments. Finally, we describe and formalize the problem addressed in this paper.

\subsection{LLM Infrastructure}
Training LLMs requires an advanced infrastructure, including GPU clusters, high-speed RDMA NICs, and ample storage capacity. In particular, the demand for NVIDIA A100 GPU devices often exceeds 5,000~\cite{narayanan2021efficient}. As LLM continues to grow in size, this demand is expected to increase further. However, the development of cluster infrastructure has lagged behind rapid advances in LLM. Many existing GPU clusters, constructed several years ago, lack the capacity to accommodate more than 1,000 NVIDIA A100 GPU devices, and network connectivity between GPU clusters in different locations is often limited. Consequently, the need to reconstruct larger GPU clusters for LLM research has become a necessity.

Moreover, the RDMA technology has two implementations: InfiniBand (IB) and RDMA over Converged Ethernet (RoCE). Different GPU clusters may be equipped with different types of RDMA NIC, and these two variants are not mutually compatible. Existing training frameworks, such as Megatron-LM~\cite{shoeybi2019megatron, narayanan2021efficient} and Megatron-DeepSpeed~\cite{rasley2020deepspeed, deepspeed}, face challenges in training LLM in the heterogeneous NIC environment. Consequently, they are unable to fully harness GPU devices in the heterogeneous clusters. This limitation leads to a suboptimal utilization of GPU computing resources.


\subsection{NIC Environments} 
We consider the scenario involving a group of GPU devices engaged in collaborative LLM training. These GPUs are from clusters equipped with different types of RDMA NIC, including IB or RoCE. We summarize the following two cases within this context.

\begin{itemize}[itemsep=1pt,topsep=1pt,parsep=1pt,leftmargin=0.4cm]
\item \textit{\textbf{Case 1: Homogeneous Clusters with High-Speed Interconnects.}} These GPU devices are distributed among multiple clusters, each equipped with the same type of RDMA NIC and connected through high-speed interconnects. All GPU devices can communicate through RDMA NICs.\label{case1}
\item \textit{\textbf{Case 2: Heterogeneous Clusters without High-Speed Interconnects.}} 
These GPU devices are distributed among multiple clusters, each equipped with either the same or distinct types of RDMA NIC. Notably, there are no high-speed interconnects between these clusters. Communication via RDMA NICs is limited to GPU devices within the same cluster, whereas GPU devices located in different clusters communicate using Ethernet NICs.\label{case2}
\end{itemize}

\subsection{Problem Description}
As illustrated in Figure \ref{problem description}, this paper focuses on scheduling computation tasklets in the heterogeneous NIC environment with varying bandwidth values. The distributed LLM training task is implemented using data and model parallelism techniques, where the entire computational task is decomposed into many parallel groups. Each group represents a set of parallel computations, with each computation viewed as a tasklet. Each tasklet corresponds to the forward and backward computations in a training iteration and is executed on certain GPU devices.

In the heterogeneous NIC environment, the training procedure is constrained by communication. The objective of Holmes is to achieve acceleration by allocating tasklets that require high communication volumes to GPU devices with faster connections.
This is achieved by leveraging the characteristics of the network environment, ensuring the maximization of device throughput in the LLM training process. The described method incorporates two parallelism techniques: pipeline and data. Pipeline parallelism involves the concurrent processing of computations in multiple stages. Each device is responsible for activation or gradient computation pertaining to different micro-batches in parallel, and the results can subsequently be communicated or transferred to subsequent stages. In addition, data parallelism implies that devices independently compute gradients for various micro-batches, necessitating synchronization of these gradients through communication.

We use the following two metrics.
\begin{itemize}[itemsep=1pt,topsep=1pt,parsep=1pt,leftmargin=0.4cm]
    \item \textbf{TFLOPS} stands for the achieved teraFLOP/s per GPU, representing the number of floating-point operations a GPU can perform in one second. It measures the GPU utilization rate, and the computational formula aligns with that in~\cite{narayanan2021efficient}. 
    \item \textbf{Throughput} stands for the number of samples processed per second $(samples/s)$ during the LLM training procedure. It measures the end-to-end training speed.
\end{itemize}

\subsection{Formalization}
The scheduling problem is formalized as follows.
\begin{itemize}[itemsep=1pt,topsep=1pt,parsep=1pt,leftmargin=0.4cm]
\item Let's take into account a collection of $M$ clusters, which we denote as $\mathbf{C} = \{c_1, c_2, \ldots,c_i,\ldots, c_M\}$. It is assumed that each cluster $c_i$ comprises $f_i$ nodes and that each node is equipped with a constant number of $G$ devices.
The total number of GPU devices is denoted as $N = G \cdot \sum_{i=1}^{M}f_i$.
\item We sequentially number clusters, nodes, and GPU devices. In the $i$-th cluster ($0 < i \leq M$), the $j$-th GPU device ($0 < j \leq G$) in the $k$-th node ($0 < k \leq f_i$) is denoted as $\operatorname{rank}_{G \cdot ((\sum_{a=1}^{i-1}f_a) + k - 1) + j}$ in the global context.
\item The pipeline parallel degree is represented by $p\ (0 < p 
\leq N/G)$, the tensor parallel degree by $t\ (0 < t 
\leq G)$, and the data parallel degree by $d\ (0 < d 
\leq N)$. It is important to note that the product of these degrees, $d \cdot p \cdot t$, is equal to $N$, which represents the total number of GPU devices.
\item For pipeline parallelism with a degree of $p$, there are a total of $t \cdot d$ pipeline parallel groups. Each of these groups can be represented as a matrix $[\textbf{PP}] \in \mathbb{Z}_{+}^{(t\ \cdot\ d) \times p}$. 
Similarly, for tensor and data parallelism with degrees $t$ and $p$, there are a total of $p \cdot d$ tensor parallel groups and $p \cdot t$ data parallel groups, respectively. These two types of group can be represented as matrices $[\textbf{TP}]\in \mathbb{Z}_{+}^{(p\ \cdot\ d) \times t}$ and $[\textbf{DP}] \in \mathbb{Z}_{+}^{(p\ \cdot\ t) \times d}$. 
\end{itemize}

Different scheduling methods may lead to different parallel groups, each with its different communication costs. Our goal is to identify the most efficient scheduling method that minimizes the communication costs of the overall training procedure in the heterogeneous NIC environment.

\section{Holmes Design}
\begin{figure*}
  \centering
  \includegraphics[width = 0.82\textwidth]{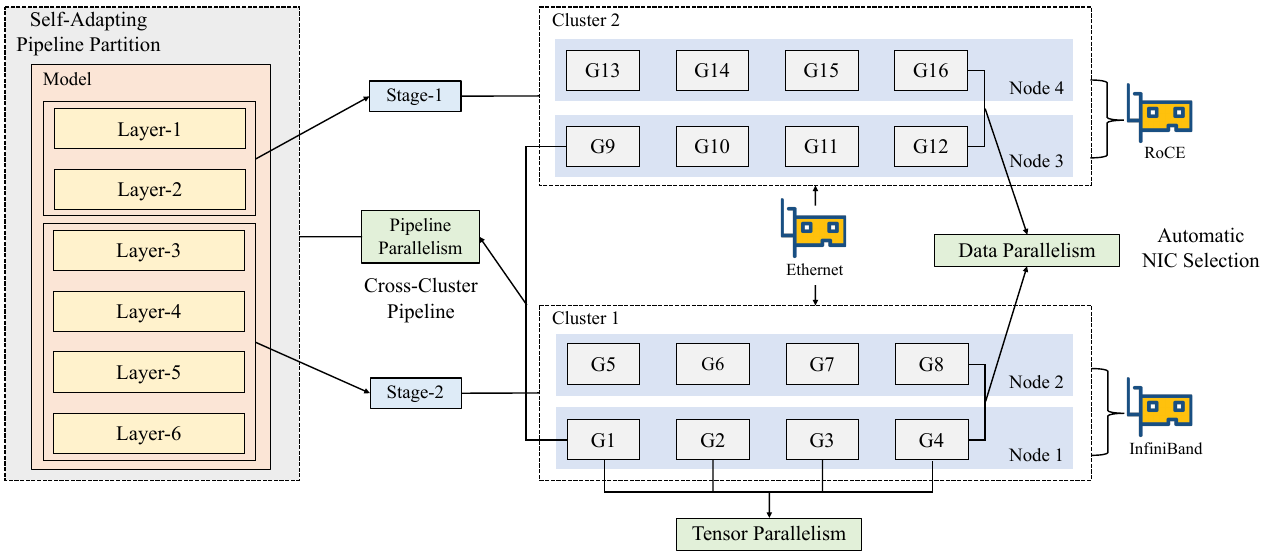}
  \caption{An illustration of parallelism mechanism in Holmes. In this case, a transformer-based model with 6 layers is trained across 2 clusters. Each cluster comprises 2 nodes, each equipped with 4 GPU devices. Communication between Node 1 and Node 2 utilizes InfiniBand, while Node 3 and Node 4 use RoCE. However, there is no high-speed interconnect between the two clusters, and communication between them relies solely on low-speed Ethernet. For parallelism settings, the degrees of data, tensor, and pipeline parallelisms are 2, 2, and 4, respectively. Pipeline parallelism is implemented between the two clusters using Ethernet. The model's layers are unevenly partitioned into 2 stages and further distributed to different GPU devices. Data parallelism is performed within each cluster using RDMA, and tensor parallelism is implemented within each node using NVLink.}
  \label{example}
\end{figure*}

In this section, we present the Holmes design. Holmes is developed within the framework of Megatron-LM~\cite{shoeybi2019megatron,narayanan2021efficient}, a popular tool for LLM training. Our approach re-optimizes two types of parallelism techniques: model parallelism and data parallelism. Model parallelism can be further categorized into tensor parallelism and pipeline parallelism. Holmes enhances the built-in communication methods for pipeline and data parallelism in Megatron-LM through optimizations including \textit{Cross-cluster Pipeline Parallelism}, \textit{Automatic NIC Selection} and \textit{Self-Adapting Pipeline Partition}. Figure~\ref{example} illustrates an overall design of the parallelism mechanism in Holmes. Further details will be discussed in the subsequent parts.

\subsection{Cross-cluster Pipeline Parallelism}
\subsubsection{Parallelism Strategy Design}
In order to facilitate a large-scale training job across multiple clusters, a dedicated parallelism strategy is required. As depicted in Figure~\ref{example}, the basic parallelism techniques could be categorized into data parallelism, pipeline parallelism, and tensor parallelism. Each parallelism technique utilizes different communication patterns and therefore exhibits a specific network requirement. Hence, the strategies of various parallelism techniques should be coherent to the heterogeneous NIC environments.

For \textbf{tensor parallelism}, it involves distributing individual layers of the model across multiple GPU devices. The computation paragidm within tensor parallelism is the operation on splited model weights and complete activation tensors. Substantial broadcast and gather operations on activation tensors are therefore implemented, resulting in substantial overhead in communication.

In the tensor parallelism of our framework, different parts of the model's layers are assigned to various GPU devices within the same node, enabling concurrent processing of these layers, and thereby accelerating the training process. As a result, the communication operations associated with tensor parallelism predominantly take place within the same node, leveraging technologies such as PCI-E~\cite{kaldewey2012gpu, li2019evaluating} or NVLink~\cite{foley2017ultra} to facilitate inter-device communication.
Holmes adopts the same partitioning strategy used by Megatron-LM for transformer layers, which serve as the fundamental components of LLMs~\cite{narayanan2021efficient}.

For \textbf{pipeline parallelism}, the model layers are segmented into different computational stages, each assigned to different GPU devices. These devices execute the stages sequentially, transferring their results to subsequent stages. Send and receive communication operations occur through the exchange of data between these devices. The communication workload of pipeline parallelism is comparably low among the three parallelism techniques. Moreover, it is possible to overlap the computation and communication between parallel stages. Therefore, the network bottleneck between clusters could be concealed by carefully designed pipeline parallelism strategy.

The implementation of our pipeline parallelism is similar to PipeDream-Flush~\cite{narayanan2021memory}. We use periodic pipeline flushes to maintain the synchronization of optimizer steps between GPU devices. Current LLM training frameworks exclusively support LLM training with low-speed Ethernet NICs within clusters that possess heterogeneous NICs. 
To address the associated limitations, we have developed a solution to facilitate LLM training in heterogeneous NIC environment. Capitalizing on the relatively low communication cost of pipeline parallelism than tensor parallelism and data parallelism, we establish pipeline parallelism groups among multiple clusters that lack high-speed interconnects. Consequently, these pipeline parallelism groups consist of GPU devices connected via heterogeneous NICs, and their sole means of communication is through Ethernet due to NIC incompatibility. Conversely, owing to the division of pipeline parallelism groups, the data parallelism groups can comprise GPU devices connected via homogeneous NICs, enabling communication via RDMA NICs. In terms of tensor parallelism, it is confined to the same node and communicates via PCI-E or NVLink interfaces.
\begin{table}
    \begin{center}
    \resizebox{0.45\textwidth}{!}{
    \begin{tabular}{lccc}
    \hline
    \textbf{NIC Env} & \textbf{TFLOPS} & \textbf{Throughtput} & \textbf{Bandwidth (Gbps)} \\ \hline
    InfiniBand & 197 & 99.23 & 200  \\ 
    RoCE       & 160 & 80.54 & 200  \\ 
    Ethernet   & 122 & 61.32 & 25   \\ \hline
    \end{tabular}
    }
    \caption{Comparison of TFLOPS and Throughput metrics when training a GPT model with 3.6 billion parameters on 4 nodes using InfiniBand, RoCE, and Ethernet respectively. Each node is equipped with 8 NVIDIA A100 GPUs. We also list the bandwidth of different NICs in the 3rd column.}
    \label{t1}
    \end{center}
\end{table}


For \textbf{data parallelism}, each GPU device possesses a complete replica of the model, while the input dataset is partitioned among these GPU devices. As previously noted, the communication cost associated with data parallelism is considerable. Communication operations in data parallelism occur in the steps including data partition, gradient computation, gradient aggregation and model update.
Communication between GPUs primarily occurs during the gradient aggregation step using all-reduce operations, facilitated by high-speed RDMA NIC and libraries like NCCL~\cite{nccl}, ensuring the synchronization of gradients between GPUs and maintaining consistent model parameters across all devices. 

Similarly to pipeline parallelism, traditional data parallelism implementation does not take into account the heterogeneous NIC environments, and it can only use the low-speed Ethernet NIC when training LLMs in the heterogeneous environment. To address the problem and accelerate the data parallelism in both homogeneous and heterogeneous environment, we introduce the following enhancement to data parallelism: an optimal selection strategy of NICs, which helps creating data parallelism groups based on the actual network environment, ensuring that GPU devices within the same data parallelism group are connected to the same type of NIC, while different data parallelism groups can select different NICs. Detailed implementation is presented in the subsequent parts. Considering that data parallelism relies more on high-speed RDMA NICs than pipeline parallelism does, our Cross-Cluster Pipeline Parallelism method has potential to significantly improve LLM training performance by scheduling pipeline parallelism groups across different clusters in heterogeneous NIC environments.


\subsubsection{Formalization}
In order to implement the aforementioned design, we propose a formalized specification of the cross-cluster pipeline parallelism strategy in terms of parallel matrices over various parallel groups of devices. 

Following the partitioning process, we 
represent $[\textbf{TP}]_{i,j}$ as the $j$-th device in the $i$-th tensor parallel group, which can be formalized as follows:
\begin{equation}
\label{eq:tp}
{[\textbf{TP}]_{i,j} = \operatorname{rank}_{(i-1) \cdot t + j}}\quad_{j \in \{1,2\ ,...,\ t\}}^{i \in \{1,2\ ,...,\ p\ \cdot\ d\}}
\end{equation}
For the $i$-th tensor parallel group $[\textbf{TP}]_{i,*}$, it comprises of $t$ devices from the same node.

To formalize the process of pipeline parallelism, we represent $[\textbf{PP}]_{i,j}$ as the $j$-th device in the $i$-th pipeline parallel group, which can be formalized as follows:
\begin{equation}
\label{eq:pp}
{[\textbf{PP}]_{i,j} = \operatorname{rank}_{i + (j-1) \cdot t \cdot d }}\quad_{j \in \{1,2\ ,...,\ p\}}^{i \in \{1,2\ ,...,\  t\ \cdot\ d\}}
\end{equation}

The formulation is applicable to both homogeneous and heterogeneous NIC environments. In the heterogeneous environment, $[\textbf{PP}]_{i,*}$ represents the $i$-th pipeline parallel group comprising $p$ GPU devices that may be connected to different types of NICs.

To formalize the process of data parallelism, we represent $[\textbf{DP}]_{i,j}$ as the $j$-th device in the $i$-th data parallel group, which can be formalized as follows:
\begin{equation}
\label{eq:dp}
{[\textbf{DP}]_{i,j} =  \operatorname{rank}_{\operatorname{mod}(i-1, t) + (\lfloor\frac{i-1}{t}\rfloor \cdot d + j - 1) \cdot t + 1}}\ _{j \in \{1,2\ ,...,\ d\}}^{i \in \{1,2\ ,...,\ p\ \cdot\ t\}}
\end{equation}
where $\operatorname{mod}(i-1, t)$ denotes the modulo operation, 
and $\lfloor\_\rfloor$ is used to signify rounding down a real number. For the $i$-th data parallel group $[\textbf{DP}]_{i,*}$, it comprises $d$ devices that are connected with the same type of NICs.

According to the aforementioned formalization, Holmes can generate the Cross-Cluster Pipeline Parallelism strategy by specifying all the parallel groups in a static manner, which further benefits the valid implementation of the parallel strategy.

\subsection{Automatic NIC Selection}
Traditional implementation of parallelism mechanisms establishes a unified communication environment for all parallel communication groups, thus making it unsuitable for GPU devices with mixed NICs. In scenarios where a data parallelism group consists of two GPU devices connected to InfiniBand and RoCE NICs, respectively, due to their inherent incompatibility, it is not possible to simultaneously select InfiniBand and RoCE NICs. Consequently, communication between the two devices is limited to Ethernet, failing to fully utilize high-speed NICs. Therefore, traditional data parallelism frequently faces performance bottlenecks, particularly during the Gradient Aggregation step, as it necessitates waiting for all GPU devices to finish their computations. We introduce an auto-select strategy by modifying NCCL~\cite{nccl} and Megatron-LM~\cite{megatronlm} to support hybrid setups of NICs, including InfiniBand, RoCE, and Ethernet. 

We first number the clusters according to the type of associated NIC. Hence, there exists a number $M_1\in \{1,2,\ldots,M\}$, such that, for all $c_i\in \mathbf{C}$, $c_i$ is equipped with IB NICs if $i\leq M_1$, otherwise, $c_i$ is equipped with ROCE NICs.
Moreover, independent communication channels are established for every parallel group according to the cross-cluster pipeline parallel strategy. 
\begin{itemize}[leftmargin=0.4cm]
    \item For every tensor parallel group $\textbf{[TP]}_{i,*}$, where $i\in\{1,2,\ldots, p\cdot d\}$, a communication group within nodes is created.
    \item For every pipeline parallel group $[\textbf{PP}]_{i,*}$, where $i\in\{1,2,\ldots, t\cdot d\}$, a communication group between pipeline stages is created using Ethernet.
    \item  For every data parallel group $[\textbf{DP}]_{i,*}$, where $i\in\{1,2,\ldots, p\cdot t\}$, a communication group is created ether using IB or ROCE, depending on the configuration of the cluster $c_j$ it belongs to.  If $j\leq M_1$, the communication group of $[\textbf{DP}]_{i,*}$ selects IB, otherwise, it selects ROCE.
\end{itemize} 

\begin{table*}[ht]
\centering
\resizebox{0.82\textwidth}{!}{
\begin{tabular}{c|c|c|c|c|c|c}
\hline
\textbf{\begin{tabular}[c]{@{}c@{}}Parameter Group\end{tabular}} & \textbf{\begin{tabular}[c]{@{}c@{}}Number of \\ Parameters (billion)\end{tabular}} & \textbf{\begin{tabular}[c]{@{}c@{}}Attention\\ Heads\end{tabular}} & \textbf{\begin{tabular}[c]{@{}c@{}}Hidden\\ Size\end{tabular}} & \textbf{\begin{tabular}[c]{@{}c@{}}Number \\ of Layers\end{tabular}} & \textbf{\begin{tabular}[c]{@{}c@{}}Pipeline \\ Parallel Size\end{tabular}} & \textbf{\begin{tabular}[c]{@{}c@{}}Batch\\ Size\end{tabular}} \\ \hline
1 & \multirow{2}{*}{3.6} & \multirow{4}{*}{32} & \multirow{2}{*}{3072} & \multirow{2}{*}{30} & \multirow{2}{*}{2} & 768 \\ \cline{1-1} \cline{7-7} 
2 &  &  &  &  &  & 1536 \\ \cline{1-2} \cline{4-7} 
3 & \multirow{2}{*}{7.5} &  & \multirow{2}{*}{4096} & \multirow{2}{*}{36} & \multirow{2}{*}{3} & 1536 \\ \cline{1-1} \cline{7-7} 
4 &  &  &  &  &  & 2688 \\ \hline
\end{tabular}%
}
  	\caption{Details of parameter groups for training GPT models ranging from 3.6 billion to 7.5 billion parameters. Since our optimization focuses on data parallelism and pipeline parallelism, without involving tensor parallelism, for Parameter Group 1 to Parameter Group 4, we set the tensor parallel size to 1.}
	\label{t2_small}
\end{table*}
\subsection{Self-Adapting Pipeline Partition}
Traditional pipeline parallelism was initially designed for training LLM in the homogeneous NIC environment. In such settings, the model layers are uniformly divided into different stages, ensuring relatively equal computational speeds across various stages of pipeline parallelism. This uniform partitioning is considered the most suitable choice, as all GPU devices exhibit equal computational speeds~\cite{narayanan2021efficient}. However, in the context of a heterogeneous NIC environment, different GPU devices may demonstrate varying computational speeds due to the diverse NICs to which they are connected. Table~\ref{t1} presents a comparison of GPU performance using three different types of NIC during LLM training. Even if InfiniBand and RoCE NICs have the same bandwidth, the GPU device equipped with different types of NIC may exhibit significant variations in actual computational speed (TFLOPS). The observed disparities in TFLOPS and Throughput metrics underscore the inadequacy of a straightforward uniform partition for pipeline parallelism in the heterogeneous NIC environment.

In response to this challenge, we have developed a self-adaptive pipeline partitioning strategy that takes into consideration the specific NIC connected to each GPU device. This strategy optimizes the training process in the heterogeneous NIC environment by distributing varying numbers of layers in the model to different devices based on their connected NIC. To formalize the Self-Adapting Pipeline Partition strategy, we define the computational speed of a device within InfiniBand and RoCE as $S(\textsc{IB})$ and $S(\textsc{RoCE})$, respectively, interpreted as TFLOPS. Assuming a pipeline parallelism group with two devices connected to different NICs and a total number of layers for the model denoted as $N$, the number of layers distributed to the device connected to the InfiniBand NIC (denoted as $N_{ib}$) is calculated as follows:
\begin{equation}
\label{eq:self-pp}
N_{ib} = \lfloor \frac{\alpha \cdot S(\textsc{IB})}{S(\textsc{IB}) + S(\textsc{RoCE})} \cdot N \rfloor 
\end{equation}
where $\alpha$ is the hyper-parameter. $N_{roce}$ can be directly obtained by subtracting $N_{ib}$ from $N$.It can be easily extended to scenarios where the pipeline parallel size is more than 2. Our objective is to allocate a greater number of model layers to the GPU device connected to the faster NIC. In addition, we also take into account a broader scenario where multiple clusters exist. The layer allocation for each cluster can be determined according to the following formula:
\begin{equation}
{ N_{c_{i}}=\lfloor \frac{\alpha_{c_{i}} \cdot S\left(c_{i}\right)}{\sum_{i}^{M} S\left(c_{i}\right)} \cdot N\rfloor \quad_{i\in \{1,2\ ,\ldots,\ M-1\}}}
\end{equation}
In total, there are $M$ clusters, where $c_i$ represents the i-th cluster. The allocation of layers for M-1 clusters needs to be adjusted by $\alpha_{c_{i}}$, which is a variable related to communication and memory. The desired $N_{c_i}$ needs to satisfy $\operatorname{Mem}\left(N_{c_{i}}\right) \leq \operatorname{DMem} \left(c_{i}\right),\ i\in \{1,2\ ,\ldots,\ M\}$. The parameter of $\operatorname{Mem}(N_{c_i})$ represents the memory occupied by cluster $c_i$ according to the current allocation scheme, and $\operatorname{DMem} \left(c_{i}\right)$ represents the maximum memory of cluster $c_i$. Through adjusting $\alpha_{c_{i}}$, we can further fine-tune the allocation scheme for pipeline parallelism.

\section{Experiment}

In this section, we seek to answer the following questions:

\textit{\textbf{Q1}}: How well does Cross-Cluster Pipeline Parallelism perform for models of different sizes?

\textit{\textbf{Q2}}: How well does Automatic NIC Selection perform in various network contexts?

\textit{\textbf{Q3}}: How well does Self-Adapting Pipeline Partition perform compared with Uniform Pipeline Partition?

\textit{\textbf{Q4}}: How well does Holmes perform compared to other mainstream LLM training frameworks? 

\textit{\textbf{Q5}}: How well does Holmes perform on different number of nodes?

\textit{\textbf{Q6}}: How well does the different component of Holmes contribute to LLM training?

\begin{table*}[h]
\begin{center}
\resizebox{0.8\textwidth}{!}{
\begin{tabular}{cccccccc}
\hline
\multirow{3}{*}{\textbf{\begin{tabular}[c]{@{}c@{}}Parameter \\ Group\end{tabular}}} & \multirow{3}{*}{\textbf{NIC Env}} & \multicolumn{2}{c}{\textbf{4 Nodes with 32 GPUs}} & \multicolumn{2}{c}{\textbf{6 Nodes with 48 GPUs}} & \multicolumn{2}{c}{\textbf{8 Nodes with 64 GPUs}} \\ \cline{3-8} 
 &  & \multicolumn{1}{c}{\multirow{2}{*}{\textbf{TFLOPS}}} & \multirow{2}{*}{\textbf{\begin{tabular}[c]{@{}c@{}}Throughput \end{tabular}}} & \multicolumn{1}{c}{\multirow{2}{*}{\textbf{TFLOPS}}} & \multirow{2}{*}{\textbf{\begin{tabular}[c]{@{}c@{}}Throughput \end{tabular}}} & \multicolumn{1}{c}{\multirow{2}{*}{\textbf{TFLOPS}}} & \multirow{2}{*}{\textbf{\begin{tabular}[c]{@{}c@{}}Throughput \end{tabular}}} \\
 &  & \multicolumn{1}{c}{} &  & \multicolumn{1}{c}{} &  & \multicolumn{1}{c}{} &  \\ \hline
\multirow{4}{*}{1} & InfiniBand & \multicolumn{1}{c}{197} & 99.23 & \multicolumn{1}{c}{188} & 142.09 & \multicolumn{1}{c}{148} & 148.88 \\ 
& RoCE & \multicolumn{1}{c}{160} & 80.54 & \multicolumn{1}{c}{151} & 114.15 & \multicolumn{1}{c}{145} & 145.64 \\
 & Ethernet & \multicolumn{1}{c}{122} & 61.32 & \multicolumn{1}{c}{99} & 74.98 & \multicolumn{1}{c}{83} & 83.38 \\
 & Hybird & \multicolumn{1}{c}{\textbf{149}} & \textbf{74.91} & \multicolumn{1}{c}{\textbf{129}} & \textbf{97.84} & \multicolumn{1}{c}{\textbf{112}} & \textbf{112.46} \\ \hline
\multirow{4}{*}{2} & InfiniBand & \multicolumn{1}{c}{206} & 103.66 & \multicolumn{1}{c}{200} & 151.25 & \multicolumn{1}{c}{156} & 156.66 \\
& RoCE & \multicolumn{1}{c}{168} & 84.78 & \multicolumn{1}{c}{162} & 122.53 & \multicolumn{1}{c}{159} & 160.47 \\
 & Ethernet & \multicolumn{1}{c}{145} & 72.95 & \multicolumn{1}{c}{128} & 96.75 & \multicolumn{1}{c}{114} & 114.52 \\ 
 & Hybird & \multicolumn{1}{c}{\textbf{162}} & \textbf{81.38} & \multicolumn{1}{c}{\textbf{152}} & \textbf{114.63} & \multicolumn{1}{c}{\textbf{132}} & \textbf{132.73} \\ \hline
\multirow{4}{*}{3} & InfiniBand & \multicolumn{1}{c}{229} & 55.95 & \multicolumn{1}{c}{220} & 80.64 & \multicolumn{1}{c}{189} & 92.35 \\ 
& RoCE & \multicolumn{1}{c}{196} & 48.04 & \multicolumn{1}{c}{185} & 67.84 & \multicolumn{1}{c}{185} & 90.40 \\
 & Ethernet & \multicolumn{1}{c}{168} & 41.04 & \multicolumn{1}{c}{143} & 52.91 & \multicolumn{1}{c}{132} & 64.85 \\ 
 & Hybird & \multicolumn{1}{c}{\textbf{191}} & \textbf{46.66} & \multicolumn{1}{c}{\textbf{170}} & \textbf{62.43} & \multicolumn{1}{c}{\textbf{168}} & \textbf{82.02} \\ \hline
\multirow{4}{*}{4} & InfiniBand & \multicolumn{1}{c}{233} & 57.03 & \multicolumn{1}{c}{228} & 83.61 & \multicolumn{1}{c}{196} & 95.79 \\ 
& RoCE & \multicolumn{1}{c}{201} & 49.10 & \multicolumn{1}{c}{193} & 70.88 & \multicolumn{1}{c}{194} & 94.85 \\
 & Ethernet & \multicolumn{1}{c}{180} & 44.10 & \multicolumn{1}{c}{168} & 61.59 & \multicolumn{1}{c}{158} & 77.31 \\ 
 & Hybird & \multicolumn{1}{c}{\textbf{200}} & \textbf{48.89} & \multicolumn{1}{c}{\textbf{187}} & \textbf{68.52} & \multicolumn{1}{c}{\textbf{177}} & \textbf{86.58} \\ \hline
\end{tabular}
}
\caption{Performance of different parameter groups in both homogeneous and heterogeneous NIC environments. Here the \textit{Hybird} environment contains two clusters with the same number of nodes.}
\label{t3}
\end{center}
\end{table*}

\subsection{Experiment Setup}
\textbf{Machine Environment.}
All of our results are run with mixed precision. Each cluster node is equipped with 8 NVIDIA A100 GPUs, each with 80 GB of memory~\cite{a100}, interconnected by NVLink~\cite{foley2017ultra}. The A100 GPUs achieve a peak device throughput of 312 teraFLOP/s with 16-bit precision. For the comparison of our results, we report results including the achieved teraFLOP/s per GPU (TFLOPS) and aggregate Throughput, calculated by multiplying the throughput of each individual GPU.

\textbf{NIC Environment.} To simulate different cases discussed in Section~\ref{case1}, we conducted experiments across different types of NIC environments. The characteristics of these NIC environments are summarized below.
\begin{itemize}[itemsep=1pt,topsep=1pt,parsep=1pt,leftmargin=0.4cm]
\item \textit{InfiniBand}: The cluster is equipped with InfiniBand NICs.
\item \textit{RoCE}: The cluster is equipped with RoCE NICs.
\item \textit{Ethernet}: The cluster lacks high-speed RDMA NICs and can only communicate through Ethernet NICs.
\item \textit{Hybird}: There are multiple different clusters, each with equipped with different high-speed NICs. The connection between clusters are restricted to Ethernet.
\end{itemize}

\textbf{Model Selection.}
For our experiments, we employ GPT models of different sizes. Specifically, the model must be compatible with the number of model-parallel GPUs used in the experiment. We utilize standard model architectures, such as GPT-3~\cite{brown2020language} when appropriate.

\textbf{Parameter Group.}
To demonstrate the end-to-end performance of Holmes in training GPT models of various sizes in the heterogeneous NIC environment, we define several parameter groups. Table~\ref{t2_small} provides detailed information on the parameter groups used in our experiments. We evaluate each parameter group with varying numbers of nodes under different NIC environments.

\textbf{Hyper-Parameter and Compared Frameworks.}
The experimental results of Holmes presented in Figure~\ref{t8-small} are based on the Self-Adapting Pipeline Partition strategy, with the hyperparameter $\alpha$ set to 1.05. The LLM training frameworks compared with Holmes include Megatron-LM~\cite{megatronlm} and Megatron-LLaMA~\cite{megatronllamma}. Notably, we incorporate the implementation of Overlapped Distributed Optimizer from Megatron-LM and Megatron-LLaMA as an enhancement on Holmes, and demonstrate that such enhancement is orthogonal to our proposed method.

\subsection{Performance Comparison}
\subsubsection{Different NIC Environments.}
As discussed in Section~\ref{case1}, we compare two cases in the heterogeneous NIC environments. For \textit{\textbf{Case 1}} and \textit{\textbf{Case 2}}, we evaluate Holmes across four types of NIC environments, including both homogeneous and heterogeneous high-speed interconnects, and various numbers of nodes. The results are presented in Table~\ref{t3}, providing insights to address questions \textbf{Q1} and \textbf{Q2}.

By analyzing the results in Table~\ref{t3}, it becomes evident that, with the integration of two components, Cross-Cluster Pipeline Parallelism, and Automatic NIC Selection, Holmes demonstrates the ability to efficiently train LLM in \textit{Hybrid} environment. It achieves performance levels that closely resemble those attainable in the homogeneous \textit{InfiniBand} and \textit{RoCE} environments, and significantly surpasses training efficiency in \textit{Ethernet} environment.

\begin{figure}[t]
  \centering
  \includegraphics[width = 0.5\textwidth]{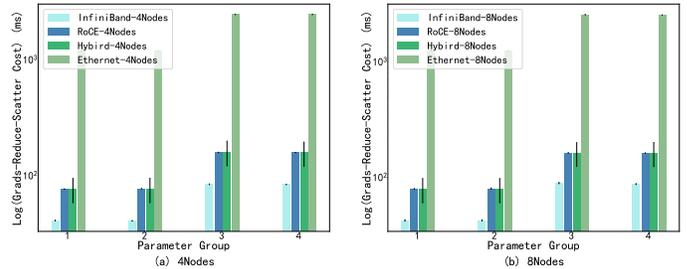}
  \caption{The time cost of \textit{grads-reduce-scatter} operation with different parameter groups in both homogeneous and heterogeneous NIC environments.}
  \label{fig:reduce-cost}
\end{figure}

\begin{figure}[t]
  \centering
  \includegraphics[width = 0.5\textwidth]{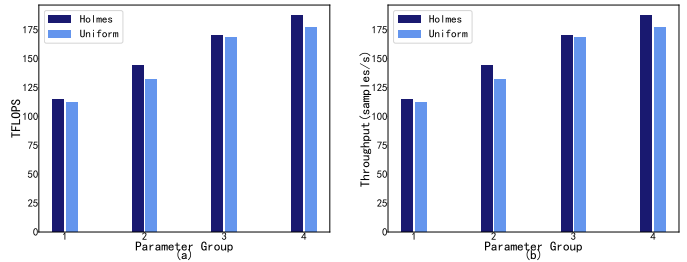}
  \caption{Performance of different parameter groups in training GPT model with various pipeline partition strategies.}
  \label{t8-small}
\end{figure}

We also present the time cost of a specific operation during LLM training in various NIC environments. Figure~\ref{fig:reduce-cost} provides a comparative result of the time cost associated with the \textit{reduce-scatter} operation, a typically time-consuming process in data parallelism. As the choice of NIC can significantly influence the time cost of this operation, the results in Figure~\ref{fig:reduce-cost} demonstrate that within the homogeneous \textit{InfiniBand} environment, \textit{reduce-scatter} exhibits the shortest duration, aligning with the results in Table~\ref{t3} where LLM training achieves the highest TFLOPS and Throughput in the homogeneous \textit{InfiniBand} environment. Furthermore, the results in \textit{Hybrid} and \textit{Ethernet} environments indicate that Holmes optimally utilizes the heterogeneous RDMA NICs, effectively reducing the time cost of \textit{reduce-scatter} and achieving faster training speed.

\subsubsection{Pipeline Partition.}
To answer question \textbf{Q3}, we conducted a comparison between Self-Adapting Pipeline Partition and Uniform Pipeline Partition. The results are presented in Figure~\ref{t8-small}. The results substantiate our hypothesis that Uniform Pipeline Partition is not the optimal choice for training LLM in the heterogeneous NIC environment. Holmes employs Self-Adapting Pipeline Partition to account for variations in training speeds between two GPU clusters, resulting in higher TFLOPS and Throughput during the LLM training process compared to traditional Uniform Pipeline Partition strategies.

\subsubsection{Comparison with Other Training Frameworks.}
\begin{figure}[t]
  \centering
  \includegraphics[width = 0.4\textwidth]{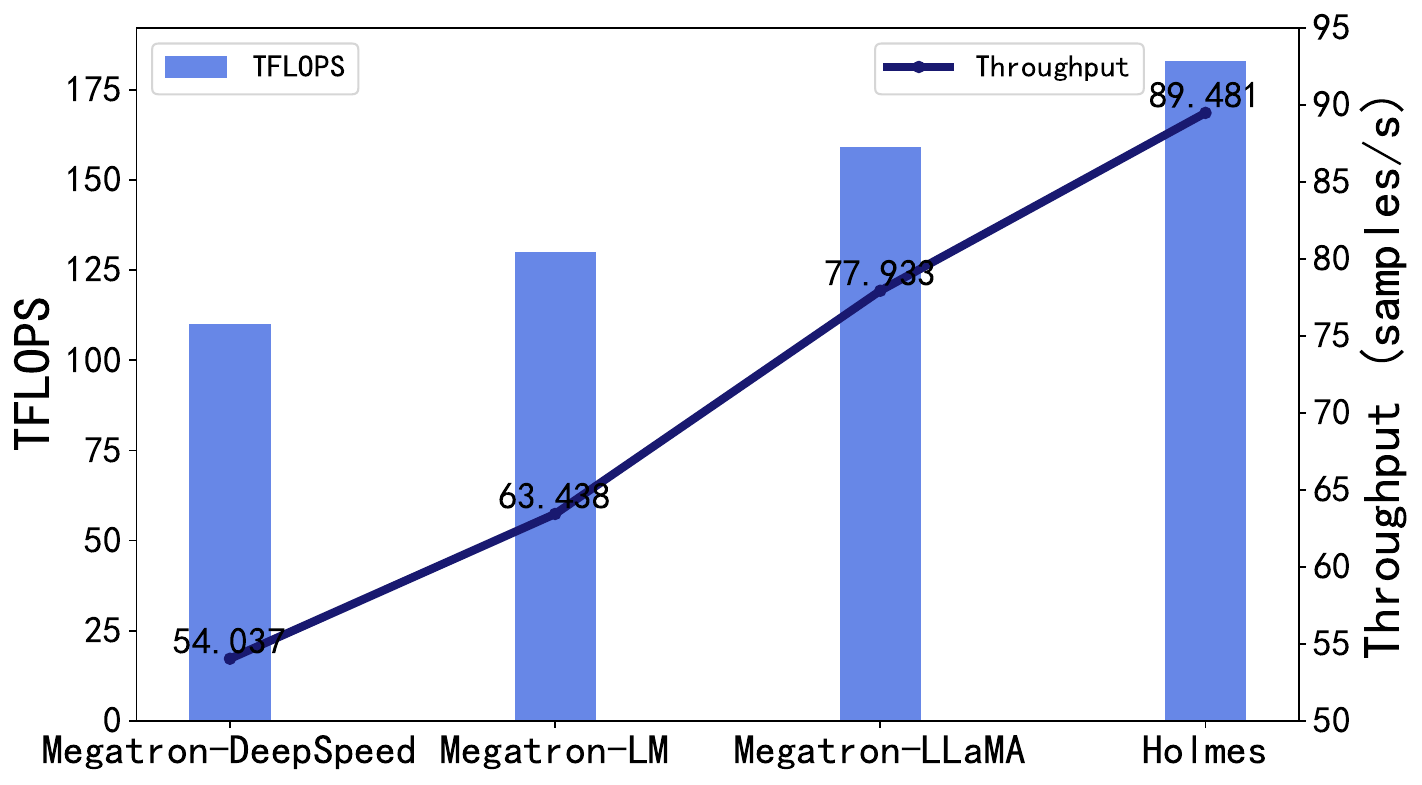}
  \caption{Performance comparison between Holmes and other mainstream LLM training frameworks: experiment conducted using parameter group 3 on 8 nodes (4 nodes equipped with RoCE NICs and 4 nodes equipped with IB NICs).}
  \label{t5}
\end{figure}

To answer question \textbf{Q4}, we conducted a comparative analysis on the performance of Holmes against other prominent LLM training frameworks, namely Megatron-LM, Megatron-DeepSpeed, and Megatron-LLaMA, within the heterogeneous NIC environment. The results of this comparison are presented in Figure~\ref{t5}. Notably, Holmes outperforms the other LLM training frameworks, emerging as the top performer. This superiority can be attributed to Holmes being the sole training framework specifically designed for heterogeneous NIC environments. 

\subsubsection{Scalability.}
\begin{figure}[t]
  \centering
  \includegraphics[width = 0.4\textwidth]{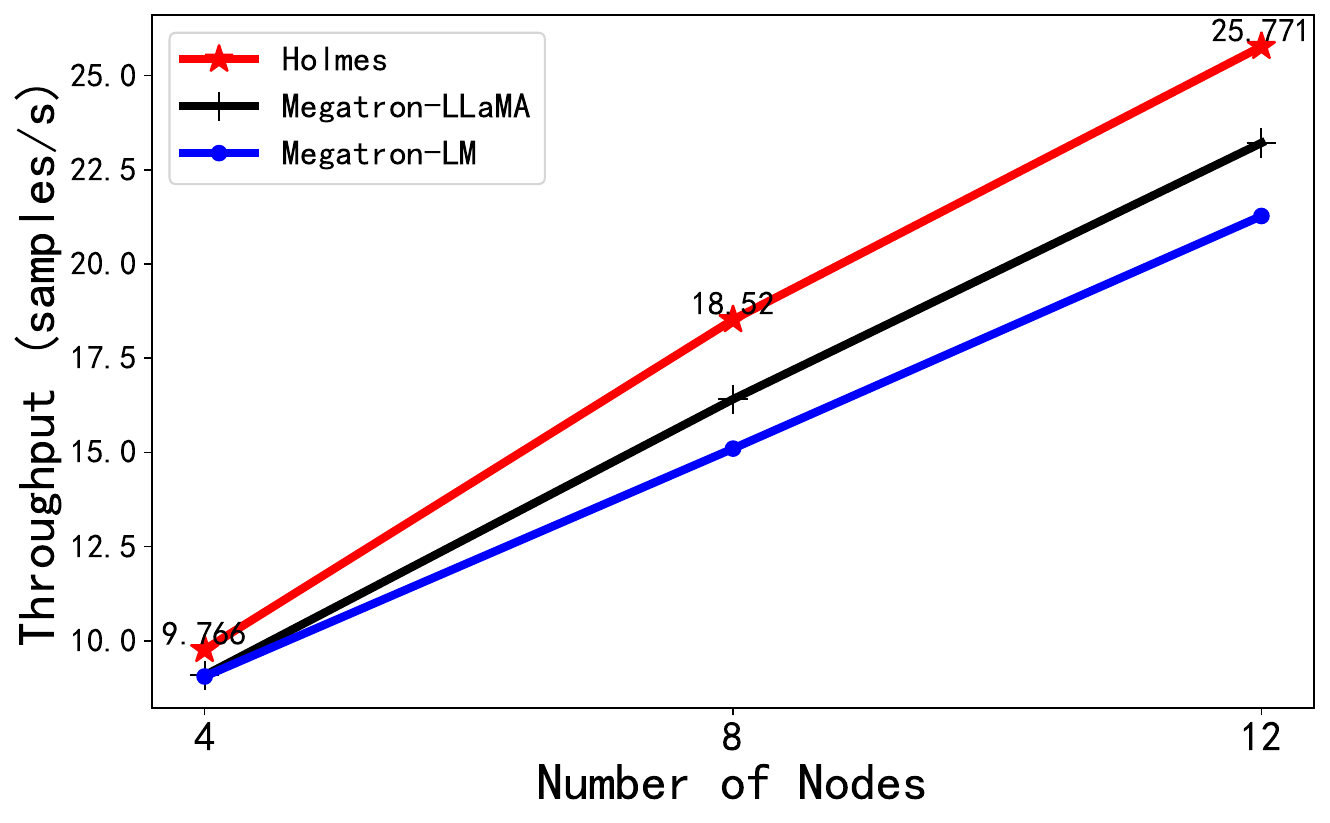}
  \caption{Comparison of speedup ratios in training GPT model with varying numbers of nodes: Holmes versus other mainstream training frameworks. The experiment was conducted on 39.1 billion parameters GPT model.}
  \label{speedup}
\end{figure}

To answer question \textbf{Q5}, we further evaluate the speedup ratios of Holmes in comparison to these frameworks as displayed in Figure~\ref{speedup}. Notebaly, Holmes achieves the highest speedup compared to other LLM training frameworks, which further demonstrates the efficiency in training scalability. We believe that Holmes can scale to distributed training involving tens to hundreds of nodes.

\subsection{Ablation Study}
The Holmes Framework comprises key components designed to enhance the training speed of LLMs in the heterogeneous NIC environment. These components include Cross-Cluster Pipeline Parallelism, Self-Adapting Pipeline Partition, Automatic NIC Selection, and Overlapped Distributed Optimizer.

To answer question \textbf{Q6} and investigate the impact of different components in Holmes, we conducted an ablation study in this subsection, as shown in Table~\ref{t6}. The effect of Cross-Cluster Pipeline Parallelism and Automatic NIC selection can be validated by a comparison between the first line and the last line of Table~\ref{t6}, also from the analysis of Table~\ref{t3} in the previous subsection. Both evidences demonstrate the effectiveness of the combination of the aforementioned components. From the remaining lines in Table~\ref{t6}, we observe that the component of Overlapped Distributed Optimizer and Self-Adapting Pipeline Partition also contributes to the training effectiveness in a nearly orthogonal way. This suggests that these two components enhance the training speed from different aspects.


\subsection{Discussion}
In this section, we conduct comprehensive experiments to address the initial questions. Firstly, we evaluate Holmes's performance across various NIC environments, a key experiment in this paper. Our findings show that Holmes competes favorably in heterogeneous NIC environments, approaching the efficiency of homogeneous RDMA-capable networks (InfiniBand or RoCE) and surpassing training efficiency in purely Ethernet environments. Next, we perform a detailed comparative analysis of Holmes's components, substantiating their effectiveness. Finally, we compare Holmes with other mainstream LLM training frameworks to validate its advancements.

\begin{table}[t]
\centering
\begin{tabular}{lcc}
\hline
\textbf{Training Framework} & \textbf{TFLOPS} & \textbf{Throughput} \\ 
\hline
Megatron-LM & 132 & 64.86 \\ \hline
Holmes & 183 & 89.48 \\ 
w/o Self-Adapting-Partition & 179 (-4) & 87.55 (-2.43) \\ 
w/o Overlapped Optimizer & 170 (-13) & 83.15 (-6.83) \\ 
w/o Above Two & 168 (-15) & 82.02 (-7.96) \\ 
\hline
\end{tabular}
\caption{Ablation study of different components in Holmes. The experiment is conducted using the same setting as Figure~\ref{t5}.}
\label{t6}
\end{table}


\section{Related Work}

\subsection{Distributed Training Frameworks}
Distributed training frameworks are essential for efficient and scalable LLM training. They enable parallel task execution across multiple devices, addressing computational and memory demands of large models. LLMs are usually trained in data centers with homogeneous high-speed RDMA NICs connecting GPUs. However, existing frameworks often neglect training LLMs in heterogeneous NIC environments.

Some popular distributed training frameworks tailored for LLMs include Megatron-LM~\cite{shoeybi2019megatron,narayanan2021efficient,megatronlm}, Megatron-LLaMA~\cite{megatronllamma}, and DeepSpeed~\cite{rajbhandari2020zero,ren2021zero,rasley2020deepspeed,deepspeed}. Megatron-LM, developed by NVIDIA, efficiently handles training of very large models using a combination of model parallelism and data parallelism across multiple GPUs and nodes. This approach enables LLM training by sharing model parameters among GPUs and processing data in parallel~\cite{narayanan2021efficient,yi2017towards}. Megatron-LM excels in minimizing synchronization delays during training through optimized communication between nodes and GPUs, utilizing efficient communication and synchronization techniques. Megatron-LLaMA, an extension of Megatron-LM, enhances the framework by integrating a standardized LLaMA implementation and introducing an efficient communication-computation parallelism method. Another notable framework is DeepSpeed, an open-source library from Microsoft designed for efficient training of large-scale deep learning models. DeepSpeed features memory-efficient training, accommodating models larger than GPU memory capacity through techniques like mixed-precision training, gradient accumulation, and offloading optimizer states. Its ZeRO optimization technology further improves memory efficiency by distributing model weights and optimizer states among GPUs and nodes, enabling training of even larger models~\cite{rajbhandari2020zero,ren2021zero}.

\subsection{Communication Optimization}
Distributed training necessitates frequent parameter synchronization. In the initial stages of communication scheduling, the primary focus is on ensuring model consistency~\cite{valiant1990bridging,recht2011hogwild,ho2013more} and refining parameter synchronization architectures~\cite{li2014scaling,patarasuk2009bandwidth,geng2018hips}. With the increasing scale of models, certain research endeavors explore coordinated optimization of both communication and parallel strategies~\cite{wang2023topoopt,zhuang2023optimizing}. This involves implementing advanced strategies to enhance parallel efficiency while concurrently reducing communication demands.

Recent advancements in distributed training efficiency, exemplified by TopoOpt~\cite{wang2023topoopt}, focus on coordinating improvements in computation, communication, and network topology. TopoOpt employs alternative optimization techniques and draws inspiration from group theory, utilizing the TotientPerms algorithm to optimize network topology, routing plans, and parallelization strategies. Another noteworthy project, Alpa~\cite{zhuang2023optimizing}, addresses challenges in cross-mesh sharding by introducing a broadcast-based communication mechanism and an overlap-friendly pipeline scheduling strategy.

\section{Conclusion}
In this paper, we present Holmes, a carefully crafted framework for LLM training across multiple GPU clusters. Our empirical studies show that this framework performs comparably to homogeneous RDMA NICs in a heterogeneous NIC environment. Holmes represents a notable advancement, enhancing accessibility to LLM training and facilitating efficient scaling within the broader research community.

\section{Acknowledgement}
This work is supported by National Natural Science Foundation of China Grant (No. U22A6001) and Key Research Project of Zhejiang Lab (No. 2022PG0AC02).

\bibliographystyle{ACM-Reference-Format}
\bibliography{sample-base}

\newpage
\section*{APPENDIX A}
\subsection*{ARTIFACT DESCRIPTION}

We present the reproducibility artifact of our experimental validation, including the description of the computational artifacts.
Our main contributions of the article lies in Holmes, an LLM training framework across GPU clusters with different type of RDMA NICs. The reproducibility artifact can assist in replicating the comparative effects mentioned in the paper with other models or methods, thereby demonstrating the efficacy of our work.

\subsubsection*{Artifact Dependencies and Requirements}
\begin{itemize}
    \item Hardware resources: XEON-GPU, NVIDIA HGX, 200G Infiniband *4 or 200G ROCE *2.
    \item Operating system: Ubuntu 18.04.4 LTS.
    \item Software libraries Dependencies: Python-3.8.10, PyTorch-2.1.0, CUDA Version: 12.1.
    \item Input dataset: we use OPT WebText dataset as input dataset to train GPT model.
    \item Other dependencies or requirements: python packages including pybind11, transformers, nltk, etc.
\end{itemize}

\subsubsection*{Artifact Installation and Deployment Process}
 
\begin{itemize}
\item Our code base is available at \textit{https://anonymous.4open.science\\/r/Holmes-40A5}. As for the runtime environment, we strongly recommend using the latest PyTorch, cuda, nccl and APEX release. You can launch an instance of the   official PyTorch image or the runnable image provided by us and mount Holmes code, your dataset with the corresponding Docker commands. And install other dependencies according to the requirement file.
\end{itemize}

 \subsubsection*{Reproducibility of Experiments}
 \begin{itemize}
    \item Our code base is capable of efficiently training large language models with both model and data parallelism over the heterogeneous NIC environment (including InfiniBand, RoCE and Ethernet). We've provided several scripts for pretraining GPT model in the examples directory of code base. To execute the script for evaluation, certain variables need to be set, including model scale(\textit{hidden size, number of heads and layers}), heterogeneous NIC environment information (\textit{NUM\_IB\_BLOCK, NUM\_GPUS\_PER\_IB\_BLOCK}), optimization training strategies switches(\textit{use-hetnet, use-asymmetric-pipeline-division, overlapped-distributed-optimizer}), and NCCL network config parameters(\textit{SOCKET\_IFNAME, HCA}).
    \item To demonstrate the framework scaling on multiple nodes and model sizes, we set variable hidden size, number of attention heads and layers to get specific model size and use up to 96 A100 GPUs (equipped with InfiniBand and RoCE NIC) to pretrain GPT models. The output results include metrics such as throughput, TFLOPS and communication operations overhead, which reflect the performance of our framework in training large language models.
    \item
    To demonstrate that Holmes reduces communication overhead during training in heterogeneous environments, we compared the \textit{Grads-Reduce-Scatter-communication time} with different parameter groups in both homogeneous and heterogeneous NIC environments.
    And through turning on the optimization switch in the script, we compare the performance of different parameter groups in training GPT model with Self-Adapting Pipeline Partition strategy and uniform partitioning method.
    Additionally, we conducted experiments on other mainstream training frameworks including Megatron-LM, Megatron-DeepSpeed, Megatron-LLaMa, and compared Holmes with these frameworks in terms of throughput, TFLOPS, and speedup rations when training the same parameterized large language models with the same number of nodes.
\end{itemize}

\end{document}